\pdfoutput=1
\documentclass[12pt]{article}
\usepackage{amsmath}
\usepackage{graphicx,psfrag,epsf}
\usepackage{enumerate}
\usepackage{natbib}
\usepackage{url} 
\usepackage{natbib}
\setcitestyle{authoryear}
\usepackage{hyperref}
\usepackage[utf8]{inputenc} 
\usepackage{amssymb}
\usepackage{graphicx}
\usepackage{graphicx} 
\usepackage{float}
\restylefloat{table}
\usepackage{placeins}
\usepackage{float}
\usepackage{wrapfig}
\usepackage{algorithm}
\usepackage{algorithmicx}
\usepackage{algpseudocode} 
\usepackage{array}
\usepackage{color}
\usepackage{csquotes}
\usepackage{longtable}
\usepackage{amsmath}
\usepackage{tikz}
\usepackage[utf8]{inputenc}
\usepackage{float}
\usepackage[skip=0pt]{caption}
\usetikzlibrary{positioning,calc}
\usetikzlibrary{shapes,fit} 

\usetikzlibrary{shapes,arrows}
\usepackage{graphicx,subfigure}

\usepackage{graphicx}
\usepackage{color}
\makeatletter
\def\cl@chapter{\@elt {theorem}}
\makeatother
\usepackage{cleveref}
\crefname{section}{§}{§§}
\Crefname{section}{§}{§§}
\definecolor{lightgray}{gray}{0.5}
\makeatletter
\newcommand{\distas}[1]{\mathbin{\overset{#1}{\kern\z@\sim}}}%
\newsavebox{\mybox}\newsavebox{\mysim}
\newcommand{\distras}[1]{%
  \savebox{\mybox}{\hbox{\kern3pt$\scriptstyle#1$\kern3pt}}%
  \savebox{\mysim}{\hbox{$\sim$}}%
  \mathbin{\overset{#1}{\kern\z@\resizebox{\wd\mybox}{\ht\mysim}{$\sim$}}}%
}
\makeatother
\def\T{{ \mathrm{\scriptscriptstyle T} }}
\newcommand{\norm}[1]{\big\Vert#1\big\Vert}

\newcolumntype{C}{>{\centering\arraybackslash}p{5em}}

\newcommand{\bx}{{\bf x}}

\newcommand{\beq}{\begin{equation}}
\newcommand{\eeq}{\end{equation}}
\newcommand{\bea}{\begin{algorithmic}}
\newcommand{\eea}{\end{algorithmic}}
\usepackage{xparse}
\NewDocumentCommand{\ceil}{s O{} m}{%
  \IfBooleanTF{#1} 
    {\left\lceil#3\right\rceil} 
    {#2\lceil#3#2\rceil} 
}
\newcommand{\blind}{1}

\newcommand{\abs}[1]{\left\vert#1\right\vert}

\begin{document}

\def\spacingset#1{\renewcommand{\baselinestretch}%
{#1}\small\normalsize} \spacingset{1}


  \title{\bf Sparse additive Gaussian process with soft interactions}
  \author{Garret Vo \hspace{.2cm}\\
    Department of Industrial and Manufacturing Engineering, \\
    Florida State University, Tallahassee, FL 32310, USA\\
    Debdeep Pati \\
    Department of Statistics,\\
    Florida State University, Tallahassee, FL 32306, USA}
  \maketitle

\if1\blind
{
  \bigskip
  \bigskip
  \bigskip
  \begin{center}
    {\LARGE\bf Title}
\end{center}
  \medskip
} \fi

\bigskip
\begin{abstract}
Additive nonparametric regression models provide an attractive tool for variable selection in 
high dimensions when the relationship between the response and predictors is complex.  They offer greater flexibility compared to parametric non-linear regression models and better interpretability and scalability than the non-parametric regression models. However, achieving sparsity simultaneously in the number of nonparametric components as well as in the variables within each nonparametric component poses a stiff computational challenge.  In this article, we develop a novel Bayesian additive regression model  using a combination of hard and soft  shrinkages to separately control the number of additive components and the variables within each component.  
An efficient algorithm is developed to select the importance variables and estimate the interaction network.  Excellent performance is obtained in simulated and real data examples.  
\end{abstract}

\noindent%
{\it Keywords:}  Additive; Gaussian process; Interaction; Lasso;  Sparsity; Variable selection 
\vfill
\hfill {\tiny technometrics tex template (do not remove)}

\newpage
\spacingset{2} 
\section{Introduction}
\indent \hspace{0.5cm} Variable selection has played a pivotal role in scientific and engineering applications, such as biochemical analysis \citep{manavalan1987variable}, bioinformatics \citep{saeys2007review}, and text mining \citep{kwon2003emotion}, among other areas.  A significant portion of existing variable selection methods  are only applicable to linear parametric models.  Despite the linearity and additivity assumption, variable selection in linear regression models has been popular since 1970; refer to  Akaike information criterion [AIC; \cite{akaike1973maximum}]; 
Bayesian information criterion [BIC; \cite{schwarz1978estimating}] and Risk inflation criterion [RIC; \cite{foster1994risk}].  

\indent\hspace{0.5cm} 
 Popular classical sparse-regression methods such as Least absolute shrinkage operator [LASSO; \cite{tibshirani1996regression,efron2004least}], and related penalization methods \citep{fan2001SCAD,Zou05regularizationand,zou2006adap,zhang2010nearly} have gained popularity over the last decade due to their simplicity, computational scalability  and efficiency in prediction when the underlying relation between the response and the predictors can be adequately described by parametric models.   Bayesian methods  \citep{mitchell1988bayesian,george1993variable,george1997approaches} with sparsity inducing priors offers greater applicability beyond parametric models and are a convenient alternative when the underlying goal is in inference and uncertainty quantification.  However,  
 there is still a limited amount of literature which seriously considers relaxing the linearity assumption, particularly when the dimension of the predictors is high.  Moreover,  when the focus is on learning the interactions between the variables, parametric models are often restrictive since they require very many parameters to capture the higher-order interaction terms.\\
\indent  \hspace{0.5cm}  Smoothing based non-additive nonparametric regression methods \citep{lafferty2008rodeo, wahba1990spline, green1993nonparametric, hastie1990generalized} can accommodate a wide range of relationships between predictors and response leading to excellent predictive performance. Such methods have been adapted for different methods of functional component selection with non-linear interaction terms: component selection and smoothing operator [COSSO; \cite{lin2006component}], sparse addictive model [SAMS; \cite{ravikumar2009sparse}] and variable selection using adaptive nonlinear interaction structure in high dimensions [VANISH; 
\cite{radchenko2010variable}]. However, when the number of variables is large and their interaction network is complex, modeling each functional component is highly expensive.\\
\indent \hspace{0.5cm} Nonparametric variable selection based on kernel methods are increasingly becoming popular over the last few years. \citet{liu2007semiparametric} provided a connection between the least square kernel machine (LKM) and the linear mixed models. \citet{zou2010nonparametric,savitsky2011variable} introduced Gaussian process with dimension-specific scalings for simultaneous variable selection and prediction. \citet{yang2015minimax} argued that a single Gaussian process with variable bandwidths can achieve optimal rate in estimation when the true number of covariates $s \asymp O(\log n)$. 
However, when the true number of covariates is relatively high, the suitability of using a single Gaussian process is questionable. Moreover, such an approach is not convenient to recover the interaction among variables. \citet{fang2012flexible} used the nonnegative Garotte kernel to  select variables and capture interaction. Though these methods can successfully perform variable selection and capture the interaction, non-additive nonparametric models are not sufficiently scalable when the dimension of the relevant predictors is even moderately high.  \cite{fang2012flexible} claimed that extensions to additive models may cause over-fitting issues in capturing the interaction between variables (i.e. capture more interacting variables than the ones which are influential). \\
\indent \hspace{0.5cm}   To circumvent this bottleneck, \citet{yang2015minimax,qamar2014additive} introduced the additive Gaussian process with sparsity inducing priors for both the number of components and variables within each component. The additive Gaussian process captures interaction among variables, can scale up to moderately high dimensions and is suitable for low sparse regression functions.  However, the use of two component sparsity inducing prior forced them to develop a tedious Markov chain Monte Carlo algorithm to sample from the posterior distribution.   In this paper, we propose a novel method, called the additive Gaussian process with soft interactions  to overcome this limitation. More specifically, we decompose the unknown regression function $F$ into $k$ components, such as $F = \sqrt{\phi_1}f_{1} + \sqrt{\phi_2} f_{2} + \dots + \sqrt{\phi_k} f_{k}$, for $k$ hard shrinkage parameters $\phi_l, l = 1, \ldots, k$, $k \geq 1$. Each component $f_{j}$ is assigned a Gaussian process prior.  To induce sparsity within each Gaussian process, we introduce an additional level of soft shrinkage parameters.  The combination of hard and soft shrinkage priors makes our approach very straightforward to implement and computationally efficient, while retaining all the advantages of the additive Gaussian process proposed by \citet{qamar2014additive}.  We propose a combination of Markov chain Monte Carlo (MCMC) and the Least Angle Regression algorithm (LARS) to select the Gaussian process components and variables within each component.  \\
\indent \hspace{0.5cm} The rest of  the paper is organized as follows. \cref{sec:method} presents the additive Gaussian process  model. \cref{sec:prior} describes the two-level regularization and the prior specifications.  The posterior computation  is detailed in \cref{sec:postcomp} and the variable selection and interaction recovery approach is presented in \cref{sec:varsel_int}.  The simulation study results are presented in  \cref{sec:sims}. A couple of real data examples are considered in \cref{sec:real}.  We conclude with a discussion in \cref{sec:conc}.    
\section{Additive Gaussian Process} \label{sec:method}
\indent \hspace{0.5cm} For observed predictor-response pairs $(\bx_i, y_i)\in \mathbb{R}^{p}\times \mathbb{R}$, where $i = 1,2,\dots, n$  (i.e. $n$ is the sample size and $p$ is the dimension of the predictors), an additive nonparametric regression model can be expressed as 
\begin{align}
\begin{split} 
y_{i} = F(\bx_i) + \epsilon_{i}, \quad \epsilon_{i} \sim \mbox{N}(0,\sigma^{2}) \\
F(\bx_{i}) =  \sqrt{\phi_1}f_{1}(\bx_{i})+ \sqrt{\phi_2} f_{2}(\bx_{i}) + \dots + \sqrt{\phi_k} f_{k}(\bx_{i}). 
\end{split}
\label{agp}
\end{align}
The regression function $F$ in \eqref{agp} is a sum of $k$ regression functions, with the relative importance of each function controlled by the set of non-negative parameters $\phi = (\phi_1, \phi_2, \ldots, \phi_k)^{\T}$.  Typically the unknown parameter $\phi$ is assumed to be sparse to prevent $F$ from over-fitting the data.  \\
\indent \hspace{0.5cm} Gaussian process (GP) \citep{rasmussen2006gaussian} provides a flexible prior for each of the component functions in $\{f_j, j=1, \ldots, k\}$. GP defines a prior on the space of all continuous  functions, denoted $f \sim \mbox{GP}(\mu, c)$ for a fixed function  $\mu: \mathbb{R}^p \to \mathbb{R}$ and a positive definite function $c$ defined on  $\mathbb{R}^p \times \mathbb{R}^p$ such that for any finite collection of points $\{\bx_l, l=1, \ldots, L\}$, the distribution of 
$\{f(\bx_1), \ldots, f(\bx_L)\}$ is multivariate Gaussian with mean $\{\mu(\bx_1), \ldots, \mu(\bx_L)\}$
and variance-covariance matrix $\Sigma =  \{c(\bx_l, \bx_{l'})\}_{1\leq, l, l' \leq L}$.  The choice of the covariance kernel is crucial to ensure the sample path realizations of the Gaussian process are appropriately smooth.  A squared exponential covariance kernel $c(\bx,\bx') = \exp(-\kappa\norm{\bx-\bx'}^{2})$ with an Gamma hyperprior assigned to the inverse-bandwidth 
parameter $\kappa$ ensures optimal estimation of an isotropic  regression function \citep{van2009adaptive} even when a single component function is used ($k=1$).  When the dimension of the covariates is high, it is natural to assume that the underlying regression function is not isotropic.  In that case,  \citet{bhattacharya2011anisotropic} showed that a single bandwidth parameter might be inadequate and dimension specific scalings with appropriate shrinkage priors are required to ensure that the posterior distribution can adapt to the unknown dimensionality.  However, \citet{yang2015minimax}  showed that single Gaussian process might not be appropriate to capture interacting variables and also does not scale well with the true dimension of the predictor space.  In that case, an additive Gaussian process is  a more effective alternative  which also leads to interaction recovery as a bi-product.  In this article, we work with the additive representation in \eqref{agp} with dimension specific scalings (inverse-bandwidth parameters) $\kappa_{lj}$ along dimension $j$ for the $l$th Gaussian process component,  $j=1, \ldots, p$ and  $l = 1, \ldots, k$. \\
\indent \hspace{0.5cm} We assume that the response vector $y = (y_{1}, y_{2}, \dots, y_{n})$ in (\ref{agp}) is 
centered and scaled. Let  $f_l \sim \mbox{GP}(0, c_l)$ with  
\begin{align}
c_l(\bx, \bx') =  \exp \bigg\{ - \sum_{j=1}^p \kappa_{lj} (x_{j} - x'_{j})^2 \bigg\}.  
\end{align} 
 In the next section, we discuss appropriate regularization on $\phi$ and  $\{\kappa_{lj}, l =1, \ldots, k; j=1, \ldots, p\}$.  A shrinkage prior on the $\{\kappa_{lj}, j=1, \ldots, p\}$ facilitates the selection of variables within component $l$ and allows adaptive local smoothing.  
An appropriate regularization on $\phi$ allows $F$ to adapt to the degree of additivity in the data without over-fitting.    
\section{Regularization} \label{sec:prior}
\indent \hspace{0.5cm} A full Bayesian specification will require placing prior distribution on both $\phi$ and 
$\kappa$.  However, such a specification requires tedious posterior sampling algorithms to sample from the posterior distribution  as seen in 
\cite{qamar2014additive}. Moreover,  it is difficult to identify the role of $\phi_l$ and $\kappa_{jl}, j=1,\ldots, p$ since one can remove the effect of the $l$ th component by  either 
setting $\phi_l$ to zero or by having  $\kappa_{jl} =0, j=1,\ldots, p$.  This ambiguous representation causes mixing issues in a full-blown MCMC.    To facilitate computation, we adopt a partial Bayesian approach to regularize $\phi$ and $\kappa_{lj}$. We propose a hybrid-algorithm which is a combination of 
i) MCMC, to sample $\kappa$ conditional on $\phi$ ii) and optimization to estimate $\phi$ conditional on $\kappa$.  With this viewpoint, we propose the following regularization on $\kappa$ and $\phi$.  
  \subsection{$L_1$ regularization for  $\phi$}  
\indent \hspace{0.5cm} Conditional on $f_1, \ldots, f_k$, \eqref{agp} is linear in $\sqrt{\phi_j}$.  Hence 
we impose  $L_1$ regularization on  $\sqrt{\phi_j}$ which are updated  using least absolute shrinkage and selection operator (LASSO)  \citep{tibshirani1996regression, tibshirani1997lasso, hastie2005elements}.  This enforces sparsity on $\phi$ at each stage of 
the algorithm, thereby pruning the un-necessary Gaussian process 
components in $F$.  
\subsection{Global-local shrinkage for $\kappa_{lj}$} \label{sparse}
\indent \hspace{0.5cm} The parameters $\kappa_{lj}$ controls the effective number of variables within each component.  For each $l$, $\{\kappa_{lj}, j=1, \ldots, p\}$ are assumed to be sparse.   
As opposed to the  two component mixture prior on $\kappa_{lj}$ in \cite{qamar2014additive}, 
we enforce weak-sparsity using a global-local continuous shrinkage prior which  potentially have substantial computational advantages over mixture priors.   
A rich variety of continuous shrinkage priors being proposed recently \citep{park2008bayesian,tipping2001sparse,griffin2010inference,carvalho2010horseshoe,carvalho2009handling,bhattacharya2014dirichlet}, which can be unified through a global-local (GL) scale mixture representation of \cite{polson2010shrink} below,
\begin{eqnarray}\label{eq:gl}
\kappa_{lj} \sim \mbox{N}(0, \psi_{lj} \tau_l), \quad \tau_l \sim f^g, \quad \psi_{lj} \sim f^l,
\end{eqnarray}
for each fixed $l$, where $f^g$ and $f^l$ are densities on the positive real line.  
In \eqref{eq:gl}, $\tau_l$ controls global shrinkage towards the origin while the local parameters $\{\psi_{lj}, j=1, \ldots,p \}$ allow local deviations in the degree of shrinkage for each predictor. Special cases include Bayesian lasso \citep{park2008bayesian}, relevance vector machine \citep{tipping2001sparse}, normal-gamma mixtures \citep{griffin2010inference} and the horseshoe \citep{carvalho2010horseshoe,carvalho2009handling} among others.  Motivated by 
the remarkable performance of horseshoe, we assume both $f^g$ and $f^l$ to be square-root of half-Cauchy distributions.  

\section{Hybrid algorithm for prediction, selection and interaction recovery} \label{sec:postcomp}
\indent \hspace{0.5cm} In this section, we develop a fast algorithm which is a combination of $L_1$ optimization and conditional MCMC to estimate the parameters $\phi_l$ and $\kappa_{lj}$  for 
$l=1, \ldots, k$.  Conditional on $\kappa_{lj}$, \eqref{agp} is linear in $\sqrt{\phi_l}$ and hence we resort to the least angle regression procedure \citep{efron2004least} with five fold cross validation to 
estimate $\phi_l, l=1, \ldots, k$.  The computation of the lasso solutions is a quadratic programming problem, and can be tackled by standard numerical analysis algorithms.\\ 
\indent \hspace{0.5cm} Next, we  describe the conditional MCMC to sample from $\kappa_{lj}$ and $F(\bx^*)$ at a new point $\bx^*$ conditional on the parameters $\phi_l$.  For two collection of vectors $X_v$ and $Y_v$ of size $m_1$ and $m_2$ respectively, denote by  $c(X_v, Y_v)$ the $m_1 \times m_2$ matrix $\{c(x, y)\}_{x \in  X_v, y \in Y_v}$.  Let ${\bf X}= \{\bx_1, \bx_2, \ldots, \bx_n\}$ and define $c({\bf X}, {\bf X}), c(\bx^*, {\bf X}), 
c({\bf X}, \bx^*)$ and $c(\bx^*, \bx^*)$ denote the  corresponding matrices.  For a random variable $q$, we denote by $q \mid -$ the conditional distribution of $q$ given the remaining random variables. \\
\indent \hspace{0.5cm} 
Observe that the algorithm does not necessarily produces samples which are approximately distributed as the true posterior distribution. The combination of optimization and conditional sampling is similar to stochastic EM \citep{diebolt1994stochastic,meng1994global} which is employed to avoid computing costly integrals required to find maximum likelihood in latent variable models.  One can expect convergence of our algorithm to the true parameters by appealing to the theory of consistency and asymptotic normality of stochastic EM algorithms \citep{nielsen2000stochastic}.  

\subsection{MCMC to sample $\kappa_{lj}$}
\indent \hspace{0.5cm} Conditional on $\phi_l, l=1, \ldots, k$, we cycle through the following steps:  
\begin{enumerate}
\item Compute $f_l^{-}(\bx_i)= \sum_{j \neq l}\sqrt{\phi_j} f_{j}(\bx_i).$
 Compute the predictive mean 
\beq
\mu_l^{*}= k(\bx^{*},\bx)[c({\bf X}, {\bf X})+\sigma^{2}I]^{-1}(y-f_l^{-})
\eeq 
\item Compute the predictive variance 
\beq
\Sigma_l^{*} = c(\bx^*, \bx^*)-c(\bx^*, {\bf X})[c({\bf X}, {\bf X})+ \sigma^{2}]^{-1}c({\bf X}, \bx^*).
\eeq
\item Sample  $f_{l}\mid -, y \sim \mbox{N}(\mu_l^{*}, \Sigma_l^{*})$.  
\item Compute the predictive 
\beq
F(\bx^*)= \sqrt{\phi_{1}}f_{1}^{*}+ \sqrt{\phi_{2}}f_{2}^{*}+\dots+ \sqrt{\phi_{k}}f_{k}^{*}.
\eeq
\item Update $\kappa_{lj}$ by sampling from the distribution 
\beq
p(\kappa_{lj}\mid -, y) \propto \frac{\exp\{-\frac{1}{2}y^{\T}[c({\bf X}, {\bf X})+\sigma^{2}I]^{-1}y}{\sqrt{|c({\bf X}, {\bf X})+\sigma^{2}I|}}p(\kappa_{lj}).
\eeq
\item Update $\tau_{l}, j=1, \ldots, k$ by sampling from the distribution
\beq
p(\tau_{l}\mid -, y)    \propto \frac{\exp\{-\frac{1}{2}y^{\T}[c({\bf X}, {\bf X})+\sigma^{2}I]^{-1}y}{\sqrt{|c({\bf X}, {\bf X})+\sigma^{2}I|}}p(\tau_l).
\eeq
\end{enumerate}

\subsubsection{Algorithm to Sample $\tau_{j}$ and $\psi_{lj}$}
\indent \hspace{0.5cm} In the MCMC algorithm above,  the conditional distributions of $\tau_{j}$ and $\kappa_{lj}$ are not available in closed form.  Therefore, we sample them using  Metropolis-Hastings algorithm \citep{hastings1970monte}. In this paper, we give the algorithm  for updating $\tau_{l}$ only, as the steps for $\kappa_{lj}$ are similar. Assuming that the chain is currently at the iteration $t$, the Metropolis-Hastings algorithm to sample $\tau_{l}^{t+1}$ independently for $l =1, \ldots, k$ proceeds as following:
\begin{enumerate}
\item Propose $\log \tau_l^{*} \sim N(\log \tau_l^{t} , \sigma_\tau^2)$. 
\item Compute the Metropolis ratio:  
\beq
p = \min \bigg[ \frac{p(\tau_l^{*} \mid - )}{p(\tau_l^{t} \mid -) }, 1  \bigg]
\eeq
\item Sample $u \sim \mbox{U}(0,1)$.
If $u < p$ then $\log \tau_l^{t+1} = \log \tau_l^{*}$, else $\log \tau_l^{t+1} = \log \tau_l^{t}$.  
\end{enumerate}    
The proposal variance $\sigma_\tau^2$ is tuned to ensure that the acceptance probability is between 20\% - 40\%.  We also propose a similar Metropolis-Hastings algorithm to sample from the conditional distribution of $\kappa_{lj} \mid -$.  

\section{Variable Selection and Interaction Recovery} \label{sec:varsel_int} 
\indent \hspace{0.5cm} In this section, we first state a generic algorithm to select important variables based on the samples of  a parameter vector $\gamma$. This algorithm is independent of the prior for $\gamma$ and unlike other variable selection algorithms, it requires little tuning parameters making  it suitable for practical purposes.  The idea is based on finding the most probable set of variables in the posterior median of $\gamma$ on $F$. Since the distribution for the number of important variables is more stable and largely unaffected by the Metropolis-Hastings algorithm, we find the mode $\mbox{H}$ of the distribution for the number of important variables. Then, we select the $\mbox{H}$ largest coefficients from the posterior mean of $\gamma$. \\
\indent \hspace{0.5cm} In this algorithm, we use $k$-means algorithm \citep{bishop2006pattern, han2011data} with $k=2$ at each iteration to form two clusters, corresponding to signal and noise variables respectively. One cluster contains values concentrating around zero, corresponding to the noise variables. The other cluster contains values concentrating away from zeros,  corresponding to the signals. At the $t^{th}$ iteration, the number of non-zero signals $h^{(t)}$ is estimated by the smaller cluster size out of the two clusters.  We take the mode over all the  iterations to obtain the final estimate $\mbox{H}$ for the number of non-zero signals i.e. $\mbox{H}=\mbox{mode}(h^{(t)})$.   
The $\mbox{H}$ largest entries of the posterior median of $\abs{\gamma}$ are identified as the non-zero signals. \\
%
\indent \hspace{0.5cm}  We run the algorithm for 5,000 iterations with a burn-in of 2,000 to ensure convergence.  Based on the remaining iterates,  we apply the algorithm to $\kappa_{jl}$ for each component $f_{l}$ to select important variables within each $f_{l}$ for $l=1, \ldots, k$.  Using this approach, we can select the important variables within each function.  We define the {\bf inclusion probability} of a variable as the proportion of functions (out of $k$) which contains that variable.    Next, we apply the algorithm to $\phi$ and select the important functions.  Let us denote by $A_f$ the set of active functions.   The {\bf probability of interaction} between a pair of variables is defined as the proportion of functions within $A_f$ in which  the pair appears together.  As a result, we can find the interaction between important variables with optimal number of active components.  
Observe that the inclusion probability and the probability of interaction is {\em not a functional of the posterior distribution} and is purely a property of the additive representation.  Hence, we do not require the sampler to convergence to the posterior distribution.  As illustrated in \cref{sec:sims},  these inclusion and the interaction probabilities provide an excellent representation of a variable or an interaction being present or absent in the model.  
\section{Simulation examples} \label{sec:sims}
\indent \hspace{0.5cm} In this section, we consider eight different simulation settings with 50 replicated datasets each and test the performance of our algorithm with respect to variable selection, interaction recovery, and prediction. To generate the simulated data,  we draw $x_{ij} \sim \mbox{Unif}(0,1)$, and $y_{i} \sim \mbox{N}(f(x_{i}),\sigma^2)$, where $1\leq i \leq n$, $1\leq j \leq p$ and $\sigma^2 =0.02$. Table \ref{tab2} summarize the results for the eight different datasets with different combinations of $p$ and $n$ for both non-interaction and interaction cases, respectively. 
\begin{table}[H]
\caption{Summary of interaction simulated datasets.}
\centering
\scalebox{0.8}{
\begin{tabular}{|l|l|l|l|l|l|l|l|l|l|l|}
\hline
    &   &  &  \multicolumn{2}{c|}{Equation for the Dataset}                                                                      \\ \hline
 Simulated Dataset & $n$ &  $p$  & Non-interaction Data & Interaction Data \\ \hline
1                 & 100 & 10            & $x_{1} + x_{2}^{2} + x_{3} + \epsilon$    & $x_{1} + x_{2}^{2} + x_{3} + x_{1}x_{2} + x_{2}x_{3} + x_{3} x_{1}+\epsilon$                     \\ \hline
2                 & 100 &100         &  $x_{1} + x_{2}^{2} + x_{3} + \epsilon$         & $x_{1} + x_{2}^{2} + x_{3} + x_{1}x_{2} + x_{2}x_{3} + x_{3} x_{1}+\epsilon$                     \\ \hline
3                 & 100 & 20            & $x_{1} + x_{2}^{2} + x_{3} + x_{4}^{2} + x_{5} + \epsilon$    & $x_{1} + x_{2}^{2} + x_{3} + x_{4}^{2} + x_{5} + x_{1}x_{2} + x_{2}x_{3} + x_{3} x_{4}+\epsilon$ \\ \hline
4                 & 100 & 100           & $x_{1} + x_{2}^{2} + x_{3} + x_{4}^{2} + x_{5} + \epsilon$    & $x_{1} + x_{2}^{2} + x_{3} + x_{4}^{2} + x_{5} + x_{1}x_{2} + x_{2}x_{3} + x_{3} x_{4}+\epsilon$ \\ \hline
\end{tabular}}
\label{tab2}
\end{table}
\subsection{Variable Selection}\label{vs::rec}
\indent \hspace{0.5cm} We compute the inclusion probability for each variable in each simulated dataset, then provide the bar plots as in Figures \ref{fig1}-\ref{fig4} below.
\begin{figure}[H]
\centering
   \subfigure[Non-interaction Case]{    
	\includegraphics[width=0.45\textwidth]{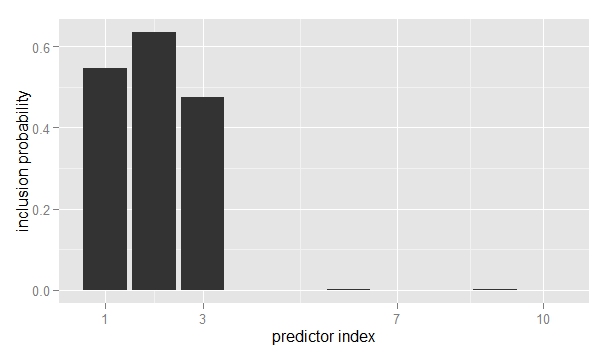}	
}
    \subfigure[Interaction Case]{    
	\includegraphics[width=0.45\textwidth]{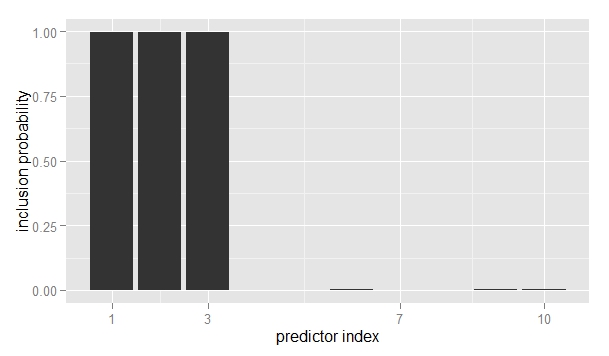}	
}   
\caption{Inclusion Probability for dataset 1.}
\label{fig1}
\end{figure}
\begin{figure}[H]
\centering
  \subfigure[Non-interaction Case]{    
	\includegraphics[width=0.45\textwidth]{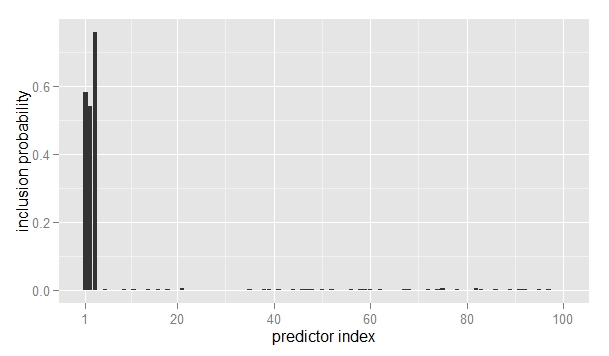}	
}
    \subfigure[Interaction Case]{    
	\includegraphics[width=0.45\textwidth]{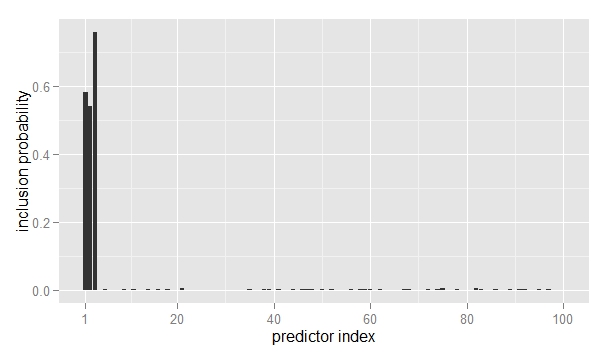}	
}   
\caption{Inclusion Probability for dataset 2.}
\end{figure}
\begin{figure}[H]
\centering
  \subfigure[Non-interaction Case]{    
	\includegraphics[width=0.45\textwidth]{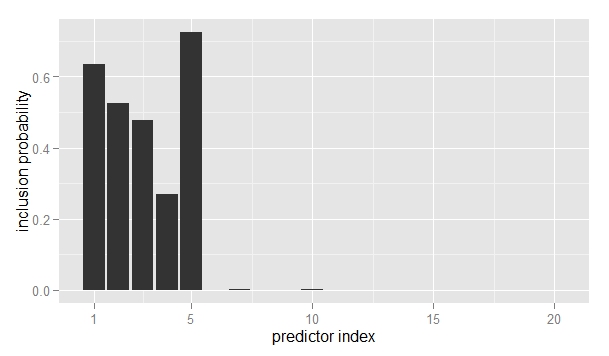}	
}
    \subfigure[Interaction Case]{    
	\includegraphics[width=0.45\textwidth]{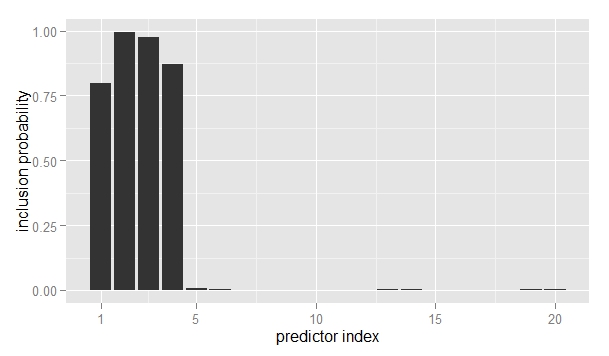}	
}   
\caption{Inclusion Probability for dataset 3.}
\end{figure}
\begin{figure}[H]
\centering
  \subfigure[Non-interaction Case]{    
	\includegraphics[width=0.45\textwidth]{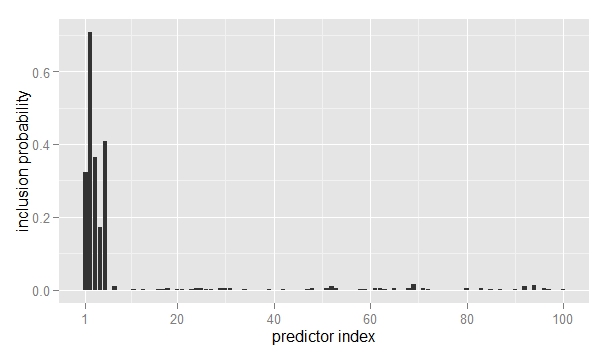}	
}
    \subfigure[Interaction Case]{    
	\includegraphics[width=0.45\textwidth]{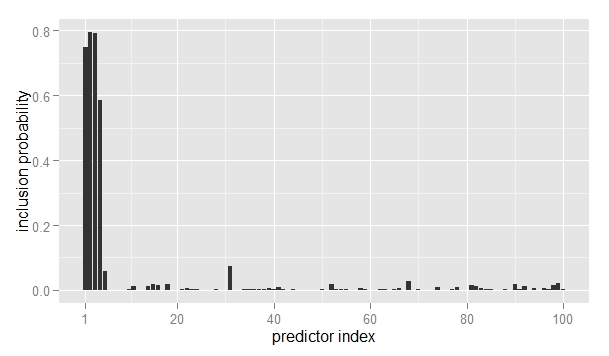}	
}   
\caption{Inclusion Probability for dataset 4.}
\label{fig4}
\end{figure}  
It is clear from the figures, that the estimated inclusion probability of the active variables are larger than the noise  variables.  To quantify the performance, we use a threshold of greater than $0.1$ to identify a variable as signal. With these included variables, we compute the false positive rate (FPR), which is the proportion of true signals not detected by our algorithm, and false negative rate (FNR), which is the proportion of false signals detected by our algorithm. Both values are recorded in Table \ref{tab3} to assess the quantitative performance of our algorithm.  
\begin{table}[H]
\centering
\caption{The average false positive (FPR) and false negative (FNR) for replicated datasets}
\begin{tabular}{|l|p{1cm}|p{1cm}|p{1cm}|p{1cm}|}
\hline
     & \multicolumn{2}{l|}{Non-interaction Dataset} & \multicolumn{2}{l|}{Interaction Dataset} \\ \hline
Dataset       & FPR             & FNR            & FPR               & FNR               \\ \hline
1            & 0.0         & 0.0        & 0.0         & 0.0     \\ \hline
2 	& 0.0  & 0.0  & 0.0 & 0.0 \\ \hline
3 	& 0.0 & 0.0 & 0.0 & 0.05\\ \hline
4 	& 0.0 & 0.01 & 0.0 & 0.01 \\ \hline
\end{tabular}
\label{tab3}
\end{table}
\vspace{-0.1cm}
Based on the results in Table \ref{tab3}, it is immediate that the algorithm is very successful in delivering accurate variable selection for both non-interaction and interaction cases. 
\subsection{Interaction Recovery}
 In order to capture the interaction network, we compute the probability of interaction between two variables by calculating the proportion of functions in which both the variables jointly appear.  With these probability values, we provide the interaction heat map 
for each dataset for both the non-interaction and interaction cases.
\begin{figure}[H]
\centering
 \subfigure[Non-interaction Case]{    
	\includegraphics[width=0.4\textwidth]{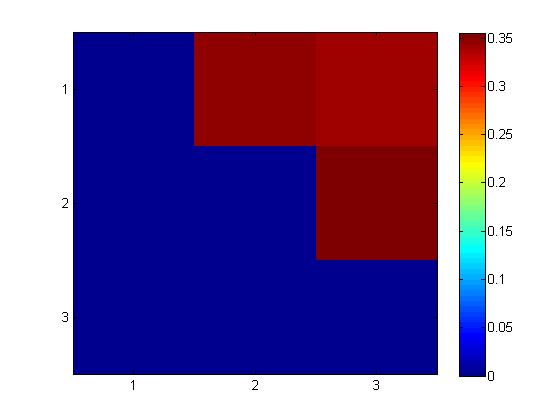}	
}
    \subfigure[Interaction Case]{    
	\includegraphics[width=0.4\textwidth]{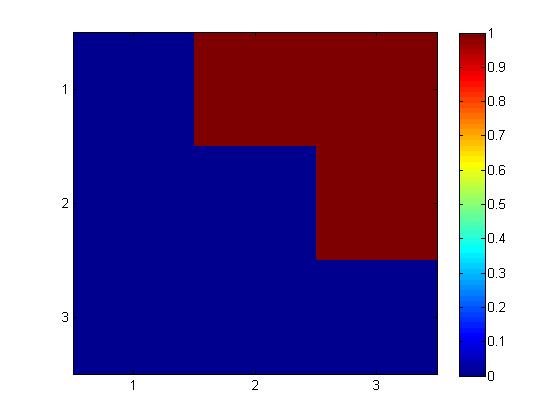}	
}   
\caption{Interaction heat map for dataset 1.}
\end{figure}
\begin{figure}[H]
\centering
 \subfigure[Non-interaction Case]{    
	\includegraphics[width=0.4\textwidth]{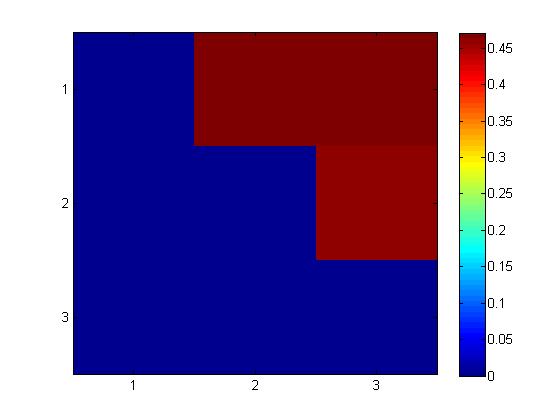}	
}
    \subfigure[Interaction Case]{    
	\includegraphics[width=0.4\textwidth]{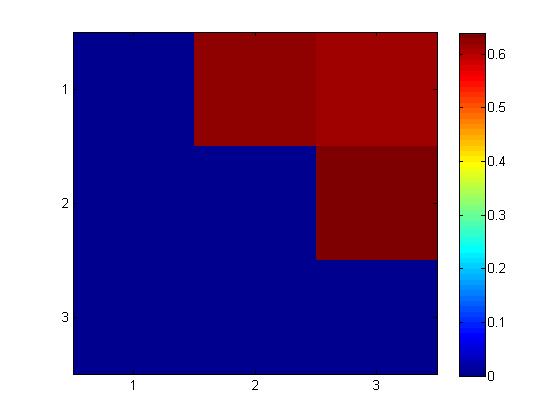}	
}   
\caption{Interaction heat map for dataset 2.}
\end{figure}
\begin{figure}[H]
\centering
 \subfigure[Non-interaction Case]{    
	\includegraphics[width=0.4\textwidth]{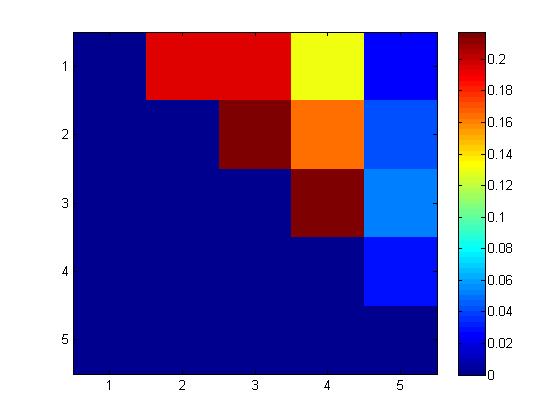}	
}
    \subfigure[Interaction Case]{    
	\includegraphics[width=0.4\textwidth]{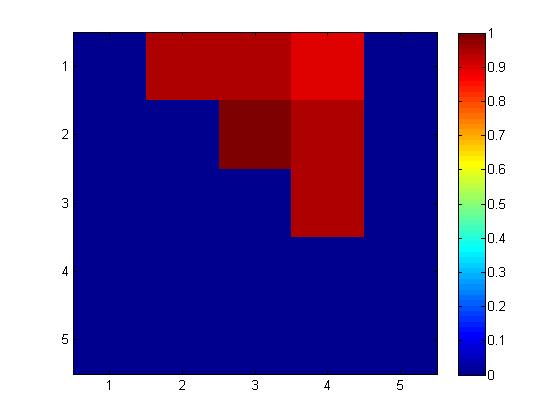}	
}   
\caption{Interaction heat map for dataset 3.}
\end{figure}
\begin{figure}[H]
\centering
 \subfigure[Non-interaction Case]{    
	\includegraphics[width=0.4\textwidth]{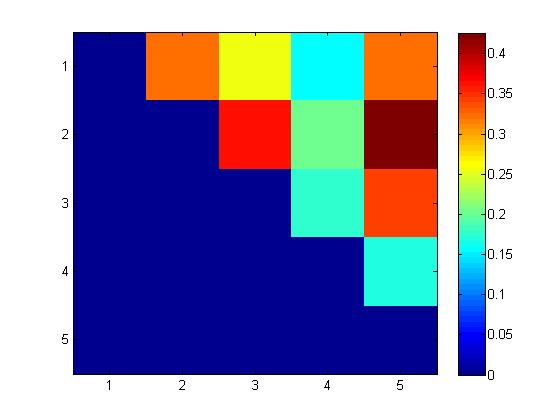}	
}
    \subfigure[Interaction Case]{    
	\includegraphics[width=0.4\textwidth]{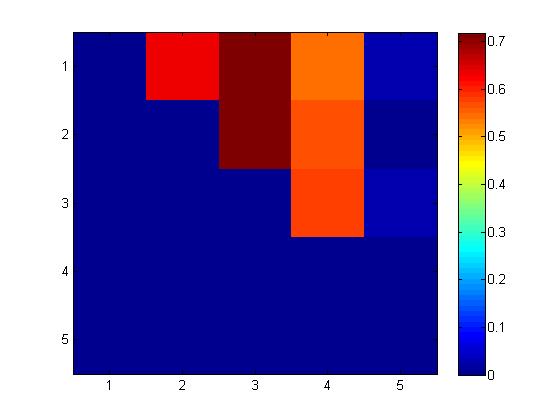}	
}   
\caption{Interaction heat map for dataset 4.}
\end{figure}  
Based on these interaction heat maps, it is evident that the estimated interaction probabilities for the non-interacting variables are less than the corresponding number for interacting variables. Using a threshold value of 0.5, we  discard these non-interacting probability values to construct  the interaction network among the variables. The interaction networks for each dataset in both non-interaction and interaction are as follows. 
\begin{figure}[H]
\centering
 \subfigure[Non-interaction 1]{    
	\includegraphics[width=0.22\textwidth]{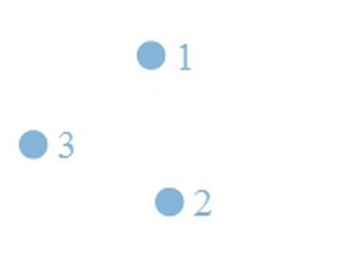}	
}
    \subfigure[Interaction 1]{    
	\includegraphics[width=0.22\textwidth]{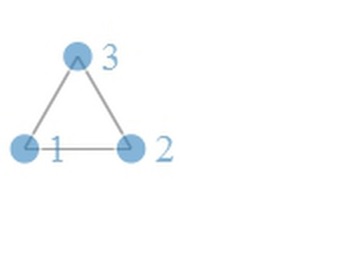}	
}   
 \subfigure[Non-interaction 2]{    
	\includegraphics[width=0.22\textwidth]{networknon1.jpg}	
}
    \subfigure[Interaction 2]{    
	\includegraphics[width=0.22\textwidth]{networkin1.jpg}	
}   %
\caption{Interaction network for dataset 1 and 2, respectively.}
\end{figure}
\begin{figure}[H]
\centering
 \subfigure[Non-interaction 3]{    
	\includegraphics[width=0.22\textwidth]{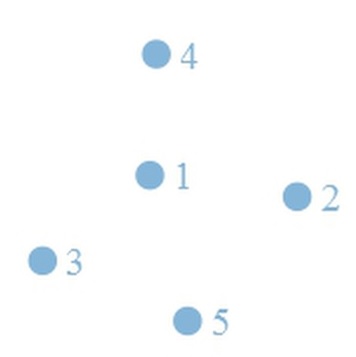}	
}
    \subfigure[Interaction 3]{    
	\includegraphics[width=0.18\textwidth]{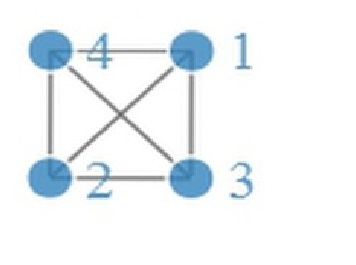}	
}   %
 \subfigure[Non-interaction 4]{    
	\includegraphics[width=0.22\textwidth]{networknon5.jpg}	
}
    \subfigure[Interaction 4]{    
	\includegraphics[width=0.22\textwidth]{network4try1.jpg}	
}   %
\caption{Interaction network for dataset 3 and 4, respectively.}
\end{figure}     
Based on the interaction network above, our algorithm successfully captures the interaction network in all the datasets for selected variables according to the inclusion probability. 
\subsection{Predictive performance}
\begin{figure}[H]
\begin{center}
 \subfigure[Prediction for Non-interaction 1]{    
	\includegraphics[width=0.45\textwidth]{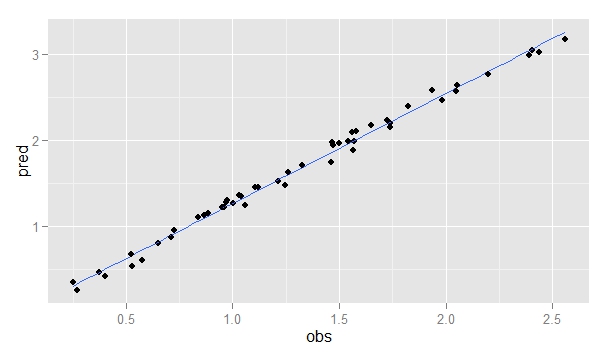}	
}
    \subfigure[Prediction for Non-interaction 3]{    
	\includegraphics[width=0.45\textwidth]{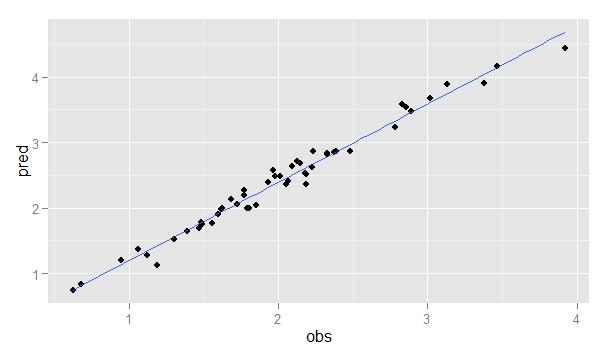}	
}   %
 \subfigure[Prediction for Interaction 1]{    
	\includegraphics[width=0.45\textwidth]{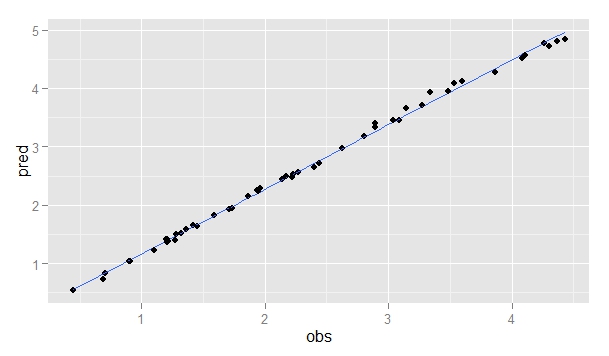}	
}
    \subfigure[Prediction for Interaction 3]{    
	\includegraphics[width=0.45\textwidth]{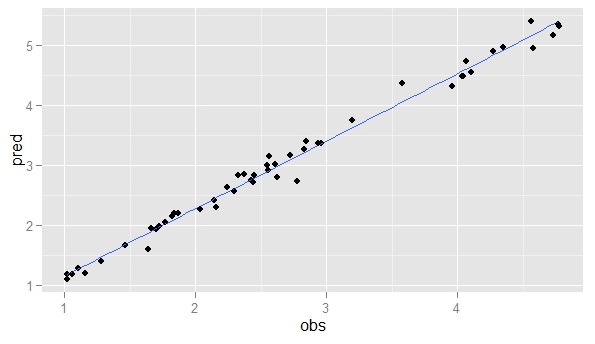}	
}   %
\caption{Prediction versus Response for Simulated Data \label{pred}}
\end{center}
\end{figure}

We randomly partition each dataset  into training (50\%) and test (50\%) observations. We apply our algorithm  on the training data and compare the performance on the test dataset.  For the sake of brevity we plot the predicted vs. the observed test observations only for a few cases  in Figure \ref{pred}.   
From Figure \ref{pred}, the predicted observations and the true observations fall very closely along the $y = x$ line demonstrating a good predictive performance.  We compare our results with \cite{fang2012flexible}. However, their additive model was not able to capture higher order interaction and  thus have a poor predictive performance compared to our method. 
\subsection{Comparison with BART} \label{compare}
Bayesian Additive Regression Tree [BART; \cite{chipman2010bart}]  is a state of the art method for variable selection in nonparametric regression problems. BART is a Bayesian ``sum of tree" framework which fits and infers the data through an iterative back-fitting MCMC algorithm to generate samples from a posterior.  Each tree in BART \cite{chipman2010bart} is constrained by a regularization prior. Hence  BART is similar to our method which also resorts to  back-fitting MCMC to generate samples from a posterior. \\
\indent \hspace{0.5cm}  Since BART is well-known to deliver excellent prediction results, its performance in terms of variable selection and interaction recovery in high-dimensional setting is worth investigating. In this section, we compare our method with BART in all the three aspects:  variable selection, interaction recovery and predictive performance.  For comparison, with BART, we  use the same simulation settings as  in Table \ref{tab2} with all combinations of ($n$, $p$), where $n=100$ and $p = 10, 20, 100, 150, 200$. \\
\indent \hspace{0.5cm} We use 50 replicated datasets and compute average inclusion probabilities for each variable. Similar to \cref{vs::rec}, the variable must have average probability value bigger than 0.1 in order to be selected. Then, we compute the false positive and false negative rates for both algorithms as in Table \ref{tab3}. These values are recorded in Table \ref{vstab}. 
\begin{table}[H]
\centering
\caption{Comparison between our algorithm and BART for variable selection}
\label{vstab}
\scalebox{0.8}{
\begin{tabular}{|l|l|l|l|l|l|l|l|l|l|l|}
\hline
        &     &     & \multicolumn{4}{c|}{Our Algorithm}          & \multicolumn{4}{c|}{BART}                 \\ \hline
Dataset & p   & n   & Non-interaction &      & Interaction &      & Non-interaction &     & Interaction &     \\ \hline
        &     &     & FPR             & FNR  & FPR         & FNR  & FPR             & FNR & FPR         & FNR \\ \hline
1       & 10  & 100 & 0.0             & 0.0  & 0.0         & 0.0  & 0.0             & 0.0 & 0.0         & 0.0 \\ \hline
2       & 100 & 100 & 0.0             & 0.0  & 0.0         & 0.0  & 1.0             & 1.0 & 1.0         & 1.0 \\ \hline
3       & 20  & 100 & 0.0             & 0.05 & 0.0         & 0.05 & 1.0             & 1.0 & 1.0         & 1.0 \\ \hline
4       & 100 & 100 & 0.0             & 0.01 & 0.0         & 0.01 & 1.0             & 1.0 & 1.0         & 1.0 \\ \hline
1       & 150 & 100 &   0.0             & 0.0     &   0.0          & 0.0      &       1.0          & 1.0    &  1.0           & 1.0    \\ \hline
4       & 150 & 100 &       0.0          & 0.0      & 0.0            &  0.0     &           1.0      & 1.0    &  1.0           &   1.0  \\ \hline
1       & 200 & 100 &     0.01            & 0.0     &       0.0      & 0.0     & NA             & NA & NA         & NA \\ \hline
4       & 200 & 100 &          0.0       &  0.0    &        0.0     & 0.0     & NA             & NA & NA         & NA \\ \hline
\end{tabular}}
\end{table}
In Table \ref{vstab}, the first column indicates which equations are used to generate the data with the respective $p$ and $n$ values in the second and third column for both non-interaction and interaction cases. For example, if the dataset is 1, the equations to generate the data is  $x_{1} + x_{2}^{2} + x_{3} + \epsilon$  and $x_{1} + x_{2}^{2} + x_{3} + x_{1}x_{2} + x_{2}x_{3} + x_{3} x_{1}+\epsilon$ for non-interaction and interaction case, respectively. NA value means that the algorithm cannot run at all for that particular combination of  $p$ and $n$ values. \\
\indent \hspace{0.5cm} According to Table \ref{vstab}, BART performs similar to our algorithm when $p=10$ and $n=100$. However, as $p$ increases, BART fails to perform adequately while our algorithm still performs well even when $p$ is less than $n$. When $p$ is twice as $n$, BART fails to run while our algorithm provides excellent results in variable selection. Overall,  our algorithm performs significantly better than BART in terms of variable selection.    
\section{Real data analysis}\label{sec:real}
In this section, we demonstrate the performance of our method in two real data sets.  We use the Boston housing data and concrete slump test datasets obtained from UCI machine learning repository. Both data have been used extensively in the literature. 
   \subsection{Boston Housing Data}
\begin{figure}[H]
\begin{center}
 \subfigure[Our Algorithm]{    
	\includegraphics[width=0.47\textwidth]{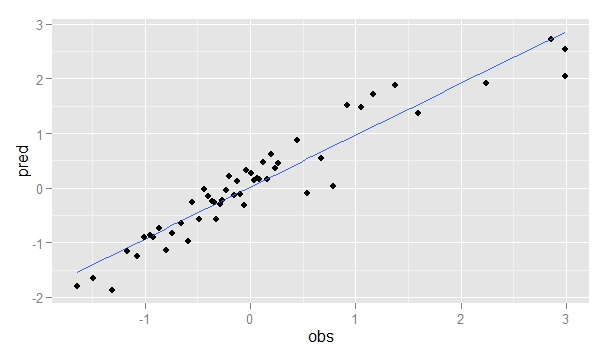}	
}
 \subfigure[BART]{    
	\includegraphics[width=0.47\textwidth]{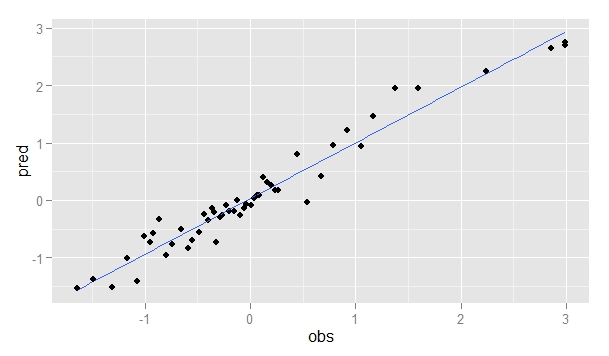}	
}
\caption{Prediction versus Response's for Boston Housing Dataset \label{predbhs}}
\end{center}
\end{figure}

\begin{figure}[H]
\begin{center}
\subfigure[Our Algorithm]{
	\includegraphics[width=0.47\textwidth]{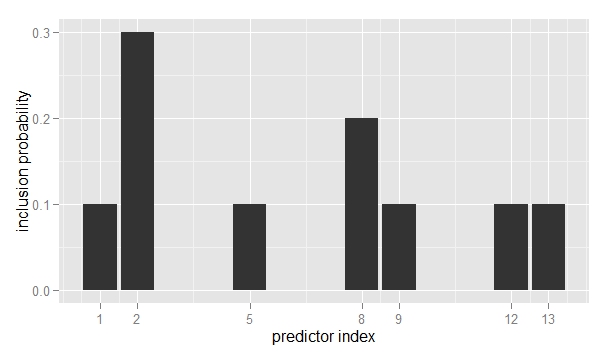}	
}
 \subfigure[BART]{    
	\includegraphics[width=0.47\textwidth]{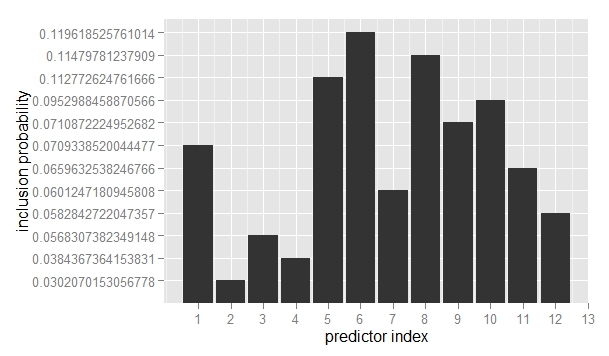}	
}
\end{center}
\caption{Inclusion Probability for the Boston Housing dataset \label{vsbhs}}
\end{figure}

\indent \hspace{0.5cm} In this section, we use the Boston housing data to compare the performance between BART and our algorithm. The Boston housing data \citep{harrison1978hedonic} contains information collected by the United States Census Service on the median value of owner occupied homes  in Boston, Massachusetts. The data has 506 number of instances with thirteen continuous variables and one binary variable. The data is split into 451 training and 51 test observations. The description for each variable is summarized in Table \ref{bhsdata}. 
 \begin{table}[H]
\centering
\caption{Boston housing dataset variable}
\label{bhsdata}
\scalebox{0.8}{
\begin{tabular}{|l|l|l|l|l|l|l|l|l|l|l|}
\hline
Variables & Abbreviation & Description   \\ \hline
 1 &  CRIM  &   Per capita crime rate  \\ \hline
  2&  ZN  &  Proportion of residential land zoned for lots over 25,000 squared feet \\ \hline
  3 & INDUS &  Proportion of non-retail business acres per town \\ \hline
   4 & CHAS  & Charles River dummy variable (= 1 if tract bounds river; 0 otherwise) \\ \hline
    5 & NOX   &  Nitric oxides concentration (parts per 10 million) \\ \hline
    6 & RM   &  Average number of rooms per dwelling \\ \hline
    7 & AGE   &    Proportion of owner-occupied units built prior to 1940 \\ \hline
    8& DIS    & Weighted distances to five Boston employment centers \\ \hline
    9 &RAD   &    Index of accessibility to radial highways \\ \hline
    10 & TAX  &    Full-value property-tax rate per \$10,000 \\ \hline
    11 & PTRATIO & Pupil-teacher ratio by town \\ \hline
    12 & B  &      $1000(\mbox{Bk} - 0.63)^{2}$ where $\mbox{Bk}$ is the proportion of blacks  by town \\ \hline
    13 &   LSTAT  & Percentage of lower status of the population \\ \hline
    14 &  MEDV  &   Median value of owner-occupied homes in \$1000's \\ \hline
\end{tabular}}
\end{table}
MEDV is chosen as the response and the remaining variables are included as predictors.  We ran our algorithm for 5000 iterations and the prediction result for both algorithms is shown in Figure \ref{predbhs}. 
Although our algorithm has a comparable prediction error with BART, we argue below that we have  a more convincing result  in terms of variable selection.    We displayed the inclusion probability barplot in Figure \ref{vsbhs}. 
 \citet{savitsky2011variable} previously analyzed this dataset and selected variables RM, DIS and LSTAT.  BART only selected 
 NOX and RM, while oue algorithm selected CRIM, ZN, NOX, DIS, B and LSTAT based on inclusion probabilities greater than or equal to 0.1.  Clearly the set of selected variables from our method has more common elements with that of  \citet{savitsky2011variable}.  

%
\subsection{Concrete Slump Test}
\begin{figure}[H]
\begin{center}
 \subfigure[Our Algorithm]{    
	\includegraphics[width=0.47\textwidth]{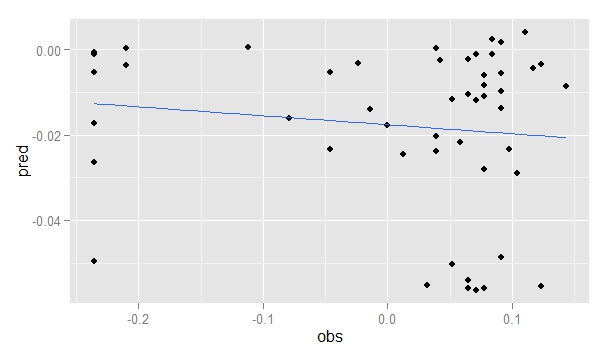}	
}
 \subfigure[BART]{    
	\includegraphics[width=0.47\textwidth]{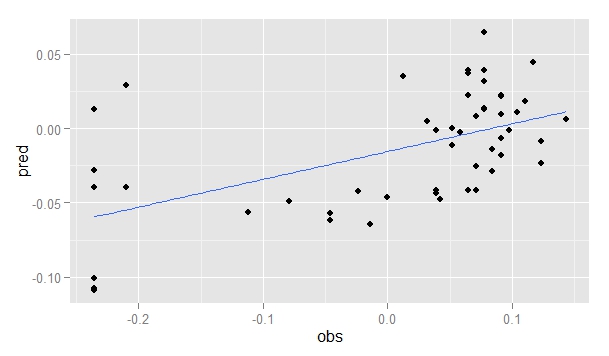}	
}
\caption{Prediction versus Response's for Concrete Slump Test Dataset \label{predslump}}
\end{center}
\end{figure}

\begin{figure}[H]
\begin{center}
\subfigure[Our Algorithm]{
	\includegraphics[width=0.47\textwidth]{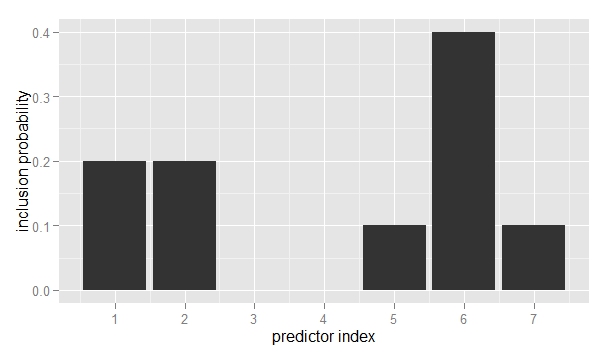}	
}
 \subfigure[BART]{    
	\includegraphics[width=0.47\textwidth]{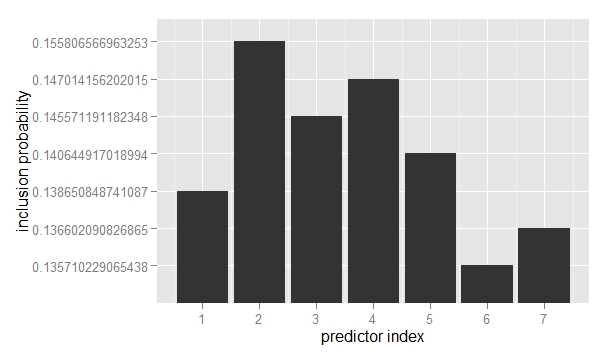}	
}
\end{center}
\caption{Inclusion Probability for the Concrete Slump Test dataset \label{hislump}}
\end{figure}

\indent \hspace{0.5cm} In this section we consider an  engineering application to compare  our algorithm against BART. The concrete slump test dataset records the test results of two executed tests on concrete to study its behavior \citep{yeh2008modeling, yeh2007modeling}.\\
\indent \hspace{0.5cm} The first test is the concrete-slump test, which measures  concrete's plasticity. Since concrete is a composite material with mixture of water, sand, rocks and cement, the first test determines whether the change in  ingredients of concrete is consistent. The first test records the change in the slump height and the flow of water. If there is a change in a slump height, the flow must be adjusted to keep the ingredients in concrete homogeneous to satisfy the structure ingenuity. The second test is the ``Compressive Strength Test", which measures the capacity of a concrete to withstand axially directed pushing forces. The second test records the compressive pressure on the concrete.  \\
\indent \hspace{0.5cm} The concrete slump test dataset has 103 instances. The data is split into 53 instances for training and 50 instances for testing. There are seven  continuous input variables, which are seven ingredients to make concrete, and three outputs, which are slump height, flow height and compressive pressure. Here we only consider the slump height as the output. The description for each variable and output is summarized in table \ref{concretedata}.  
 \begin{table}[H]
\centering
\caption{Concrete Slump Test dataset}
\label{concretedata}
\scalebox{0.8}{
\begin{tabular}{|l|l|l|l|l|l|l|l|l|l|l|}
\hline
Variables & Ingredients & Unit  \\ \hline
 1 & Cement & kg \\ \hline
2 & Slag & kg  \\ \hline
 3 & Fly ash & kg \\ \hline
4 & Water & kg \\ \hline
5 & Super-plasticizer (SP) & kg \\ \hline
6 & Coarse Aggregation & kg \\ \hline
7 & Fine Aggregation & kg  \\ \hline
8 & Slump &  cm \\ \hline
9 & Flow & cm \\ \hline
10 & 28-day Compressive Strength & Mpa \\ \hline 
\end{tabular}}
\end{table} 
The predictive performance is illustrated in figure \ref{predslump}.  
Similar to the Boston housing dataset, our algorithm performs closely to BART in prediction. Next, we investigate the performances in terms of variable selection.  We plot the barplot of the inclusion probability for each variable in Figure \ref{hislump}.  
 \citet{yurugi2000mix} determined that coarse aggregation has a significant impact on the plasticity of a concrete. Since the difference in slump's height is to measure the plasticity of a concrete, coarse aggregation is a critical variable in the concrete slump test. According to Figure \ref{hislump}, our algorithm selects coarse aggregation as the most important variable unlike BART, which clearly demonstrates the efficacy of our algorithm compared to BART.  
\section{Conclusion} \label{sec:conc}
\indent \hspace{0.5cm} In this paper, we propose a novel Bayesian nonparametric approach  for variable selection and interaction recovery with excellent performance in selection and interaction recovery in both simulated and real datasets. Our method obviates the computation bottleneck in recent unpublished work \cite{qamar2014additive} by proposing a simpler regularization involving a combination of hard and soft shrinkage parameters. Moreover, our algorithm is  computationally efficient and highly scalable.   

Although such sparse additive models are well known to adapt to the underlying true dimension of the covariates \citep{yang2015minimax}, literature on consistent selection and interaction recovery in the context of nonparametric regression models is missing. 
As a future work, we propose to investigate consistency of the variable selection and interaction
of our method.  
\bibliographystyle{spbasic} 
\bibliography{publication2}

\begin{thebibliography}{51}
\providecommand{\natexlab}[1]{#1}
\providecommand{\url}[1]{{#1}}
\providecommand{\urlprefix}{URL }
\expandafter\ifx\csname urlstyle\endcsname\relax
  \providecommand{\doi}[1]{DOI~\discretionary{}{}{}#1}\else
  \providecommand{\doi}{DOI~\discretionary{}{}{}\begingroup
  \urlstyle{rm}\Url}\fi
\providecommand{\eprint}[2][]{\url{#2}}

\bibitem[{Akaike(1973)}]{akaike1973maximum}
Akaike H (1973) Maximum likelihood identification of gaussian autoregressive
  moving average models. Biometrika 60(2):255--265

\bibitem[{Bhattacharya et~al(2014{\natexlab{a}})Bhattacharya, Pati, and
  Dunson}]{bhattacharya2011anisotropic}
Bhattacharya A, Pati D, Dunson D (2014{\natexlab{a}}) Anisotropic function
  estimation using multi-bandwidth {G}aussian processes. The Annals of
  Statistics 42(1):352--381

\bibitem[{Bhattacharya et~al(2014{\natexlab{b}})Bhattacharya, Pati, Pillai, and
  Dunson}]{bhattacharya2014dirichlet}
Bhattacharya A, Pati D, Pillai NS, Dunson DB (2014{\natexlab{b}})
  Dirichlet-laplace priors for optimal shrinkage. Journal of the American
  Statistical Association (just-accepted):00--00

\bibitem[{Bishop(2006)}]{bishop2006pattern}
Bishop CM (2006) Pattern recognition and machine learning. springer

\bibitem[{Carvalho et~al(2009)Carvalho, Polson, and
  Scott}]{carvalho2009handling}
Carvalho C, Polson N, Scott J (2009) Handling sparsity via the horseshoe. In:
  International Conference on Artificial Intelligence and Statistics, pp 73--80

\bibitem[{Carvalho et~al(2010)Carvalho, Polson, and
  Scott}]{carvalho2010horseshoe}
Carvalho CM, Polson NG, Scott JG (2010) The horseshoe estimator for sparse
  signals. Biometrika p asq017

\bibitem[{Chipman et~al(2010)Chipman, George, and McCulloch}]{chipman2010bart}
Chipman HA, George EI, McCulloch RE (2010) Bart: Bayesian additive regression
  trees. The Annals of Applied Statistics pp 266--298

\bibitem[{Diebolt et~al(1994)Diebolt, Ip, and Olkin}]{diebolt1994stochastic}
Diebolt J, Ip E, Olkin I (1994) A stochastic em algorithm for approximating the
  maximum likelihood estimate. Tech. rep., Technical Report 301, Department of
  Statistics, Stanford University, California

\bibitem[{Efron et~al(2004)Efron, Hastie, Johnstone, and
  Tibshirani}]{efron2004least}
Efron B, Hastie T, Johnstone I, Tibshirani R (2004) Least angle regression. The
  Annals of statistics 32(2):407--499

\bibitem[{Fan and Li(2001)}]{fan2001SCAD}
Fan J, Li R (2001) Variable selection via nonconcave penalized likelihood and
  its oracle properties. JASA 96(456):1348--1360,
  \doi{10.1198/016214501753382273},
  \eprint{http://dx.doi.org/10.1198/016214501753382273}

\bibitem[{Fang et~al(2012)Fang, Kim, and Schaumont}]{fang2012flexible}
Fang Z, Kim I, Schaumont P (2012) Flexible variable selection for recovering
  sparsity in nonadditive nonparametric models. arXiv preprint arXiv:12062696

\bibitem[{Foster and George(1994)}]{foster1994risk}
Foster DP, George EI (1994) The risk inflation criterion for multiple
  regression. The Annals of Statistics pp 1947--1975

\bibitem[{George and McCulloch(1993)}]{george1993variable}
George EI, McCulloch RE (1993) Variable selection via gibbs sampling. Journal
  of the American Statistical Association 88(423):881--889

\bibitem[{George and McCulloch(1997)}]{george1997approaches}
George EI, McCulloch RE (1997) Approaches for {B}ayesian variable selection.
  Statistica sinica 7(2):339--373

\bibitem[{Green and Silverman(1993)}]{green1993nonparametric}
Green PJ, Silverman BW (1993) Nonparametric regression and generalized linear
  models: a roughness penalty approach. CRC Press

\bibitem[{Griffin and Brown(2010)}]{griffin2010inference}
Griffin J, Brown P (2010) Inference with normal-gamma prior distributions in
  regression problems. {B}ayesian Analysis 5(1):171--188

\bibitem[{Han et~al(2011)Han, Kamber, and Pei}]{han2011data}
Han J, Kamber M, Pei J (2011) Data mining: concepts and techniques: concepts
  and techniques. Elsevier

\bibitem[{Harrison and Rubinfeld(1978)}]{harrison1978hedonic}
Harrison D, Rubinfeld DL (1978) Hedonic housing prices and the demand for clean
  air. Journal of environmental economics and management 5(1):81--102

\bibitem[{Hastie et~al(2005)Hastie, Tibshirani, Friedman, and
  Franklin}]{hastie2005elements}
Hastie T, Tibshirani R, Friedman J, Franklin J (2005) The elements of
  statistical learning: data mining, inference and prediction. The Mathematical
  Intelligencer 27(2):83--85

\bibitem[{Hastie and Tibshirani(1990)}]{hastie1990generalized}
Hastie TJ, Tibshirani RJ (1990) Generalized additive models, vol~43. CRC Press

\bibitem[{Hastings(1970)}]{hastings1970monte}
Hastings WK (1970) Monte carlo sampling methods using markov chains and their
  applications. Biometrika 57(1):97--109

\bibitem[{Kwon et~al(2003)Kwon, Chan, Hao, and Lee}]{kwon2003emotion}
Kwon OW, Chan K, Hao J, Lee TW (2003) Emotion recognition by speech signals.
  In: INTERSPEECH, Citeseer

\bibitem[{Lafferty and Wasserman(2008)}]{lafferty2008rodeo}
Lafferty J, Wasserman L (2008) Rodeo: sparse, greedy nonparametric regression.
  The Annals of Statistics pp 28--63

\bibitem[{Lin et~al(2006)Lin, Zhang et~al}]{lin2006component}
Lin Y, Zhang HH, et~al (2006) Component selection and smoothing in multivariate
  nonparametric regression. The Annals of Statistics 34(5):2272--2297

\bibitem[{Liu et~al(2007)Liu, Lin, and Ghosh}]{liu2007semiparametric}
Liu D, Lin X, Ghosh D (2007) Semiparametric regression of multidimensional
  genetic pathway data: Least-squares kernel machines and linear mixed models.
  Biometrics 63(4):1079--1088

\bibitem[{Manavalan and Johnson(1987)}]{manavalan1987variable}
Manavalan P, Johnson WC (1987) Variable selection method improves the
  prediction of protein secondary structure from circular dichroism spectra.
  Analytical biochemistry 167(1):76--85

\bibitem[{Meng and Rubin(1994)}]{meng1994global}
Meng XL, Rubin DB (1994) On the global and componentwise rates of convergence
  of the em algorithm. Linear Algebra and its Applications 199:413--425

\bibitem[{Mitchell and Beauchamp(1988)}]{mitchell1988bayesian}
Mitchell TJ, Beauchamp JJ (1988) Bayesian variable selection in linear
  regression. Journal of the American Statistical Association
  83(404):1023--1032

\bibitem[{Nielsen(2000)}]{nielsen2000stochastic}
Nielsen SF (2000) The stochastic em algorithm: estimation and asymptotic
  results. Bernoulli pp 457--489

\bibitem[{Park and Casella(2008)}]{park2008bayesian}
Park T, Casella G (2008) The {B}ayesian {l}asso. Journal of the {A}merican
  Statistical Association 103(482):681--686

\bibitem[{Polson and Scott(2010)}]{polson2010shrink}
Polson N, Scott J (2010) Shrink globally, act locally: sparse {B}ayesian
  regularization and prediction. {B}ayesian Statistics 9:501--538

\bibitem[{Qamar and Tokdar(2014)}]{qamar2014additive}
Qamar S, Tokdar ST (2014) Additive gaussian process regression. arXiv preprint
  arXiv:14117009

\bibitem[{Radchenko and James(2010)}]{radchenko2010variable}
Radchenko P, James GM (2010) Variable selection using adaptive nonlinear
  interaction structures in high dimensions. Journal of the American
  Statistical Association 105(492):1541--1553

\bibitem[{Rasmussen(2006)}]{rasmussen2006gaussian}
Rasmussen CE (2006) Gaussian processes for machine learning

\bibitem[{Ravikumar et~al(2009)Ravikumar, Lafferty, Liu, and
  Wasserman}]{ravikumar2009sparse}
Ravikumar P, Lafferty J, Liu H, Wasserman L (2009) Sparse additive models.
  Journal of the Royal Statistical Society: Series B (Statistical Methodology)
  71(5):1009--1030

\bibitem[{Saeys et~al(2007)Saeys, Inza, and Larra{\~n}aga}]{saeys2007review}
Saeys Y, Inza I, Larra{\~n}aga P (2007) A review of feature selection
  techniques in bioinformatics. bioinformatics 23(19):2507--2517

\bibitem[{Savitsky et~al(2011)Savitsky, Vannucci, and
  Sha}]{savitsky2011variable}
Savitsky T, Vannucci M, Sha N (2011) Variable selection for nonparametric
  gaussian process priors: Models and computational strategies. Statistical
  science: a review journal of the Institute of Mathematical Statistics
  26(1):130

\bibitem[{Schwarz et~al(1978)}]{schwarz1978estimating}
Schwarz G, et~al (1978) Estimating the dimension of a model. The annals of
  statistics 6(2):461--464

\bibitem[{Tibshirani(1996)}]{tibshirani1996regression}
Tibshirani R (1996) Regression shrinkage and selection via the lasso. Journal
  of the Royal Statistical Society Series B (Methodological) pp 267--288

\bibitem[{Tibshirani et~al(1997)}]{tibshirani1997lasso}
Tibshirani R, et~al (1997) The lasso method for variable selection in the cox
  model. Statistics in medicine 16(4):385--395

\bibitem[{Tipping(2001)}]{tipping2001sparse}
Tipping M (2001) Sparse {B}ayesian learning and the relevance vector machine.
  The Journal of Machine Learning Research 1:211--244

\bibitem[{van~der Vaart and van Zanten(2009)}]{van2009adaptive}
van~der Vaart AW, van Zanten JH (2009) Adaptive bayesian estimation using a
  gaussian random field with inverse gamma bandwidth. The Annals of Statistics
  pp 2655--2675

\bibitem[{Wahba(1990)}]{wahba1990spline}
Wahba G (1990) Spline models for observational data, vol~59. Siam

\bibitem[{Yang et~al(2015)Yang, Tokdar et~al}]{yang2015minimax}
Yang Y, Tokdar ST, et~al (2015) Minimax-optimal nonparametric regression in
  high dimensions. The Annals of Statistics 43(2):652--674

\bibitem[{Yeh et~al(2008)}]{yeh2008modeling}
Yeh I, et~al (2008) Modeling slump of concrete with fly ash and
  superplasticizer. Computers and Concrete 5(6):559--572

\bibitem[{Yeh(2007)}]{yeh2007modeling}
Yeh IC (2007) Modeling slump flow of concrete using second-order regressions
  and artificial neural networks. Cement and Concrete Composites 29(6):474--480

\bibitem[{Yurugi et~al(2000)Yurugi, Sakata, Iwai, and Sakai}]{yurugi2000mix}
Yurugi M, Sakata N, Iwai M, Sakai G (2000) Mix proportion for highly workable
  concrete. Proceedings of Concrete pp 579--589

\bibitem[{Zhang(2010)}]{zhang2010nearly}
Zhang CH (2010) Nearly unbiased variable selection under minimax concave
  penalty. The Annals of Statistics pp 894--942

\bibitem[{Zou et~al(2010)Zou, Huang, Lee, and Hoeschele}]{zou2010nonparametric}
Zou F, Huang H, Lee S, Hoeschele I (2010) Nonparametric bayesian variable
  selection with applications to multiple quantitative trait loci mapping with
  epistasis and gene--environment interaction. Genetics 186(1):385--394

\bibitem[{Zou(2006)}]{zou2006adap}
Zou H (2006) The adaptive lasso and its oracle properties. JASA
  101(476):1418--1429, \doi{10.1198/016214506000000735},
  \eprint{http://dx.doi.org/10.1198/016214506000000735}

\bibitem[{Zou and Hastie(2005)}]{Zou05regularizationand}
Zou H, Hastie T (2005) Regularization and variable selection via the elastic
  net. Journal of the Royal Statistical Society, Series B 67:301--320

\end{thebibliography}
\end{document}